\documentclass[letterpaper, 10 pt, conference]{ieeeconf}  % Comment this line out if you need a4paper

\usepackage{amssymb}
\usepackage{amsmath}
\usepackage{graphicx} 
\usepackage{times}
\usepackage{epsfig} 
\usepackage{graphicx}
\usepackage{amsmath}
\usepackage{amssymb}
\usepackage{float}
\usepackage{comment}
\usepackage{lipsum}
\usepackage{caption}    
\usepackage{subfig}    
\usepackage[most]{tcolorbox}
\usepackage{hyperref}
\usepackage{algorithm}
\usepackage[english]{babel}
\usepackage[utf8]{inputenc}
\usepackage{algorithm}
\usepackage[noend]{algpseudocode}
\usepackage{lipsum}
\usepackage{tikz}

\IEEEoverridecommandlockouts  
\title{\LARGE \bf
CObRaSO: Compliant Omni-Direction Bendable Hybrid Rigid and Soft OmniCrawler Module 
}

\author{ Enna Sachdeva*$^{1}$, Akash Singh*$^{1}$, Vinay Rodrigues$^{1}$, Abhishek Sarkar$^{1}$, K.Madhava Krishna$^{1}$% <-this % stops a space
\thanks{$^{*}$Equal Contribution}%
\thanks{$^{1}$All authors are with Robotics Research Center, IIIT-Hyderabad,  Gachibowli-500032, India}%
\thanks{{\tt\small akashvnit2016@gmail.com}}%
\thanks{{\tt\small sachdeva.enna@research.iiit.ac.in}}%
\thanks{{\tt\small rodriguesvinay10@gmail.com }}%
\thanks{{\tt\small abhishek.sarkar@iiit.ac.in }}%
\thanks{{\tt\small mkrishna@iiit.ac.in}}%
}

\begin{document}

\maketitle
\thispagestyle{empty}
\pagestyle{empty}

%%%%%%%%%%%%%%%%%%%%%%%%%%%%%%%%%%%%%%%%%%%%%%%%%%%%%%%%%%%%%%%%%%%%%%%%%%%%%%%%
\begin{abstract}
This paper presents a novel design of an Omnidirectional bendable Omnicrawler module- CObRaSO. Along with the longitudinal crawling and sideways rolling motion, the performance of the OmniCrawler is further enhanced by the introduction of Omnidirectional bending within the module, which is the key contribution of this paper. The Omnidirectional bending is achieved by an arrangement of two independent 1-DOF joints aligned at $90^\circ$ w.r.t each other. The unique characteristic of this module is its ability to crawl in Omnidirectionally bent configuration which is achieved by a novel design of a 2-DOF roller chain and a backbone of a hybrid structure of a soft-rigid material. This hybrid structure provides compliant pathways for the lug-chain assembly to passively conform with the orientation of the module and crawl in Omnidirectional bent configuration, which makes this module one of its kind. Furthermore, we show that the unique modular design of CObRaSO unveils its versatility by achieving active compliance on an uneven surface, demonstrating its applications in different robotic platforms (an in-pipeline robot, Quadruped and snake robot) and exhibiting hybrid locomotion modes in various configurations of the robots. The mechanism and mobility characteristics of the proposed module have been verified with the aid of simulations and experiments on real robot prototype.\end{abstract}

%%%%%%%%%%%%%%%%%%%%%%%%%%%%%%%%%%%%%%%%%%%%%%%%%%%%%%%%%%%%%%%%%%%%%%%%%%%%%%%%
\section{INTRODUCTION}
Mobile robots capable of traversing unstructured environment are highly useful for inspection, exploration of the unknown and risky areas as well as for Urban search and rescue (USAR) operations. High level of adaptability on uneven rough terrain and efficient maneuvering in confined spaces are the desired key functionalities of these robots. The performance of a robot locomotion is determined by the trafficability, maneuverability, and terrainability \cite{c1}, and these pose important challenges in the design of search and rescue robots. 
%, locomotion, planning, and control of these robots and has opened new areas of research/ lead to the development of new areas of research. 
\begin{comment}
\begin{figure}[t]
\centering
\hspace{4cm}
\subfloat[]{\includegraphics[width=0.4\textwidth,height=0.14\textheight]{protoype.jpg}\label{fig:module}}\\
\hspace{1cm}
\caption{Prototype of the Proposed Module.}
\label{fig:robot_view} 
\end{figure}
\end{comment}

\begin{figure}[t]
\centering
\hspace{0cm}
\subfloat[]{\includegraphics[width=0.32\textwidth,height=0.12\textheight]{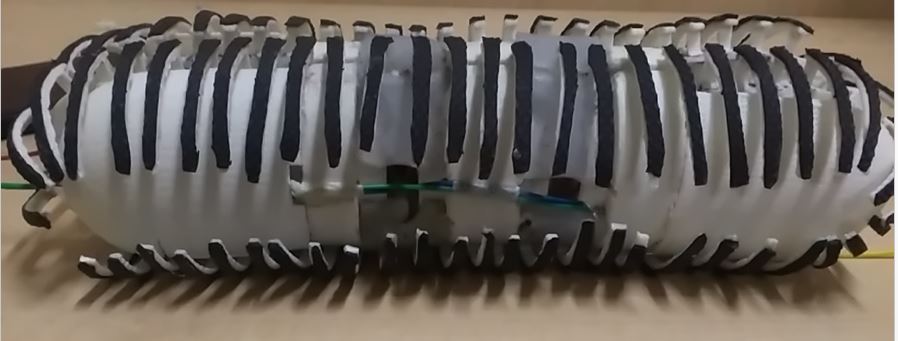}\label{fig:robot_view}}
\subfloat[]{\includegraphics[width=0.18\textwidth,height=0.12\textheight]{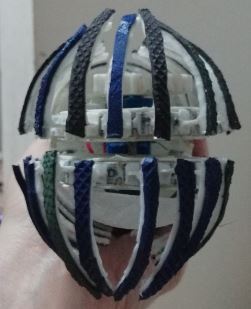}\label{fig:cross_section}}
\hspace{0cm} 
\centering
\subfloat[]
{\includegraphics[width=0.5\textwidth,height=0.08\textheight]{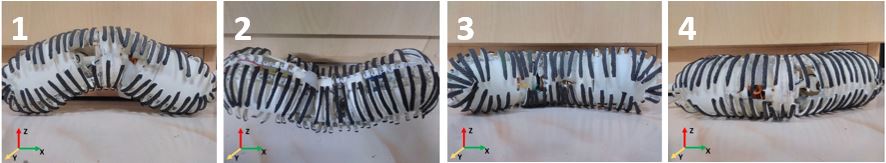}\label{fig:module_bend} }

\caption{(a) Prototype of the CObRaSO,(b) Circular Cross-section, (c) CObRaSO bending about [1-2] Y-axis (Pitch), [3-4] Z-axis (Yaw) }
\label{fig:module}
\end{figure}

%The formidable capabilities of wheeled, legged and tracked based mobile robots on uneven terrain are well documented in literature (ADD REFERENCE). Land based robots are generally based on wheeled locomotion, due to their fast rolling and energy efficient motion on hard-flat terrain \cite{c2}. Despite good maneuverability and steering capabilities, wheeled robots may fail to drive over obstacles of size larger than the wheel diameter. Moreover, omni-directional motion on uneven terrain for a wheeled robot even with mecanum or omni-wheels might not be possible \cite{c7}. Comparatively, biologically inspired legged robots offers many advantages when navigating in uneven natural terrains as they possess good adaptability by varying the effective length and orientation of the legs \cite{c2}.
%On the downside, they require more effort to control and maintain stability. 
%This has led to the development of a number of hybrid wheel-legged robots which combine the efficient motion capabilities of wheeled robots with good terrain negotiating capabilities of legged robots \cite{c2}, \cite{12} and\cite{c15}. 
%These robots can move over flat terrain using wheels and overcome complicated terrain using legs stepping. 

The impressive capabilities of wheeled, legged and tracked based mobile robots on the rugged surface are well documented in literature \cite{c17}. Land-based robots are generally based on wheeled locomotion, due to their fast rolling and energy efficient motion on hard-flat surface \cite{c2}. Legged robots possess good adaptability by varying the effective length and orientation of the legs which facilitates them to navigate in uneven natural terrains and overcome obstacles much better than wheeled robots \cite{c2}. Continuous Track robots are favorable due to their added advantages of crawling over holes and ability to smooth out the path, putting low pressure on terrain and providing large ground contact surface and traction \cite{c3}. Moreover, for high-risk missions such as military applications, where robots are required to have good traction and stability for the successful mission, tracked robots performs difficult locomotion and dexterous manipulation tasks \cite{c19}. These potential advantages of each locomotion mode have led to the development of a number of robots with hybrid locomotion modes such as wheel-leg, leg-track, wheel-track, and wheel-leg-track. Robots exhibiting hybrid locomotion modes take the potential advantages of each one while attempting to avoid their drawbacks. These robots exploit the most appropriate locomotion mode for the prevailing conditions in the environment. Francois et. al. \cite{c4} developed a multimodal wheel-leg-track locomotion robotic platform- AZIMUT, where each of the 4 independent articulated modules can generate a variety of motions such as omnidirectional motion, climbing obstacles and stairs. Zhu et. al. \cite{c5} also proposed a hybrid leg-wheel-track robot, which consists of a robot body with four driving mechanisms, four independent track devices, two supporting legs and one wheel lifting mechanism. Similarly, there are several other hybrid robots such as Transformable wheel-track robot\cite{c6}, Tri-Wheel \cite{c7}, TITAN-X\cite{c20}. 

%However, the robot can only achieve omnidirectional motion in the  achieve omnidirectional motion when each module is crawling in the longitudinal direction.

%reconfigurable robots with crawling modes along with  wheeled and legged locomotion \cite{c4}, \cite{c5}, \cite{c6} as well as

The state-of-the-art hybrid robots discussed above incorporate conventional tank-like crawler modules which can only support longitudinal (forward and backward) crawling motion. The motion capabilities and terrain adaptability of these robots on an uneven surface have been improved with the development of articulated multi-tracked robots by linking several passive or active tracked modules \cite{c8}, \cite{c12}. These robots require individual actuators for each of its articulated crawler module, which leads to an increase in size and weight of the robot. This further attenuates their capability to realize easy and efficient turning motion in constrained spaces. This paper proposes a novel design of an Omni-directional bendable OmniCrawler module- CObrRaSO, which incorporates several cascaded sub-modules with a pair of the chain going over the entire length of the module. A couple of sprocket-chain pairs is driven by a single actuator and therefore, the compact structure of the module facilitates steering and turning motion in constrained spaces with minimum actuators.

In our previous work, we have designed an OmniCrawler module with 1-DOF active compliant joints and the motion capabilities of the mechanism were demonstrated with its applications in an in-pipe climbing robot (COCrIP) \cite{c10}. The objective of the present work is to extend the capabilities and scope of applications of the OmniCrawler module in several mobile robots platforms. The Omnidirectional joints in CObrRaSO achieve compliance in the desired direction, which is the key contribution of this paper. This Omnidirectional bending capability enables it to exhibit different locomotion modes- wheel, leg and crawler in different robotic platforms. Its design is based on an OmniCrawler module, inspired by \cite{c9}. The Omnidirectional bending is characterized by the novel design of a 2-DOF compliant roller chain and a hybrid structure of soft-rigid materials. The unique design of the lugs and the ability of the chain to conform with the bent chassis of the module enables crawling motion even in the bent configuration. This property of achieving Omnidirectional bending while crawling makes this module one of its kind. Therefore, the robot can achieve turning and steering motion using a single actuator (which drives both the lug-chain assemblies). The motion of lug-chain assembly over the surface of the bent module further enables it to exhibit surface actuation property, which helps in obstacle aided motion. The surface actuation robot developed previously by McKenna \cite{c13} has demonstrated crawling motion only in the straight configuration since the toroidal skin used for locomotion seems to become loose while bending and does not maintain enough tension for crawling in a bent configuration.

%The state-of-the art hybrid robots discussed above incorporates conventional tank crawler modules. However, similar to conventional crawlers, the design of the Omnicrawler module as a single rigid body limits its capabilities to climb big steps/stairs and conform to highly unstructured environment. While the terrain adaptability and mobility of the crawler modules have been improved by developing an articulated multi-tracked robot by linking several passive or active tracked modules \cite{c18}, this leads to an increase in the number of actuators and hence the size of the robot which limits its accessibility in narrow and confined spaces. Furthermore, a number of compliant robots have been developed to dynamically comply with uneven terrain \cite{23},\cite{26}. For instance, \cite{23} develops an 'OUROBOT' robot built up of closed, deformable servo segments that changes its form to achieve high adaptability on rugged terrain. However,the inability to conform in 3 dimensions limits its applications to planar surfaces. 

The objective of this research and key contributions of this paper are summarized as below:\\
1. A seminal concept of Omnidirectional bending in an OmniCrawler module is introduced. This is realized using a hybrid combination of soft and rigid materials and a unique design of a 2-DOF roller chain. \\
2. The versatility of its modular design provides potential advantages of attaining various configurations and facilitates its integration with various robotic platforms, such as in-pipe climbing robot, Quadruped, snake robot, robotic manipulator, gripper and legged robots.\\
3. The Omnidirectional bending enhances its capabilities to exhibit various locomotion modes in various robots. For instance, with the hybrid locomotion modes, the Quadruped can navigate through tight spaces by switching its configuration between legged and crawling locomotion modes. \\

\section{Mechanical design Overview}
\label{Sec:Mechanical Design Overview}

\begin {comment}
\begin{figure}[t]
\centering
%\hspace{-2cm}
\begin{comment}
\subfloat[Exploded side view ]{\includegraphics[width=0.4\textwidth,height=0.18\textheight]{cad_bearing.PNG}}

\hspace{1cm}
\includegraphics[width=0.5\textwidth,height=0.2\textheight]{rolling_motion.jpg}
\caption{Cross-sectional view of module.}
\label{fig:cross_section} 
\end{figure}
 \end{comment}
 
\begin{figure}[t]
{\includegraphics[width=0.5\textwidth,height=0.25\textheight]{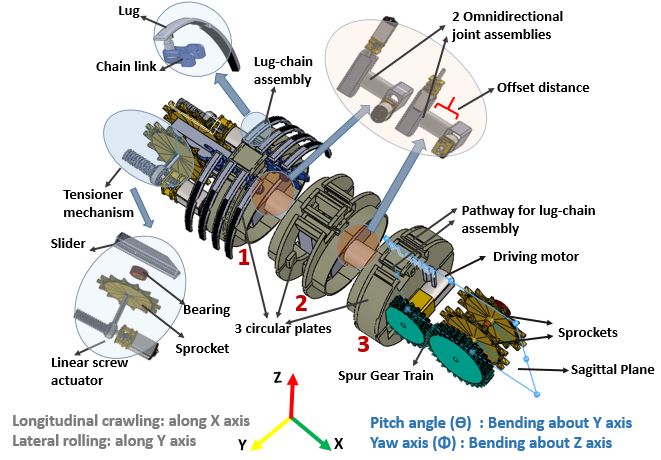}} 
\caption{ CAD model of the Proposed module  }
\label{fig:CAD}
\end{figure}

\subsection{Omnidirectional bendable OmniCrawler Module}
\label{Sec:Module}
The OmniCrawler module consists of a couple of chain-sprocket power transmission pairs on either side of the Sagittal plane of the module. Two longitudinal series of lugs resting on the chain links through attachments give a circular cross-section to the module and the ends of the module attain hemispherical shape. These two lug-chain-sprocket assemblies are driven by a micro-motor through a spur-gear train with a gear ratio of 1.25:1 (r:1). Torque driving the each of the two lug-chain assemblies $\tau_s$ can be calculated as follows.

\begin{align}
\begin{split}
 \tau_{s} = \frac{r \tau_m}{2}
\end{split}
\label{eqn:gear_ratio}
\end{align}

where $\tau_m$ is the rated torque of the geared driving motor.

\begin{table}[h!] 
\caption{Design Parameters of the module}
\label{table:Design Parameters}
\centering
\hspace{0cm}
\begin{tabular}{|l|c|r|}
\hline
\textbf{Quantity} & \textbf{Symbol} & \textbf{Values}  \\
\hline
mass of module & $M_m$ &  0.320kg  \\
\hline
length of module 
& $L_m$ &  0.24m  \\
\hline
Driving motors saturation torque & $\tau_{max}$  & 1.5Nm  \\
\hline
Joint motors saturation torque & $\tau_{max}$  & 1.5Nm  \\
\hline
Diameter of circular plate &$d$ & 0.040m \\
\hline
Joint limit of omnidirectional joint & $\theta_{max}$, $\phi_{max}$   & $32^\circ$  \\
\hline
Diameter of module & $D$ & 0.060m  \\
\hline
Thickness of silicone rubber & $t$ & 0.004m  \\
\hline
Width of silicone rubber & $w$ & 0.010m \\
\hline
Length of silicone rubber & $ls$ & 0.022m  \\
\hline
Length of lug & $l_{lug}$ & 0.010m  \\
\hline
Height of lug & $h_{lug}$ & 0.035m  \\
\hline
Separation between plates & $d$ & 0.025m  \\ 
\hline
\end{tabular}
\end{table}

\subsubsection{Omnidirectional bending}
\label{bending}

The exploded CAD view of the module demonstrating the arrangement of chains, lugs, motor mounts, bearings, gears, screw driven linear actuators and chassis in the module, is shown in Fig. \ref{fig:CAD}. The chassis of the module consists of 3 circular plates which hold an arrangement of 2 independent 1-DOF joints aligned at $90^\circ$ with respect to each other. The Omnidirectional bending is realized by controlling the rotation of these joints about Pitch and Yaw axes, as shown in Fig.\ref{fig:module_bend}. 

\subsubsection{An arrangement of two independent 1-DOF joints aligned at 90$^\circ$}

An assembly of Omnidirectional joint consists of two independent 1-DOF joints aligned at 90$^\circ$ w.r.t each other at an offset distance between them. The module consists of two such assemblies of joints held together by 3 circular plates. The central plate (Plate No.2) holds 2 motors aligned along Y and Z axes; and each of the other two circular plates holds one motor aligned individually along Y (Plate No.3) or Z (Plate No.1) axis, respectively. The arrangement of these joints in between the circular plates is shown in CAD model in Fig. \ref{fig:CAD}. The width and diameter of each of these 3 circular plates have been determined by the design constraints posed by the size of actuators incorporated inside them. 

\begin{comment}
\begin{figure}
\hspace{-0.5cm}
\begin{minipage}[t]{.12\textwidth}
\includegraphics[width=1\textwidth, height=0.12\textheight]{omni_joint.JPG}
  \captionof{figure}{ Two 1-DOF joints aligned at 90$^\circ$, with an offset distance w.r.t to each other}
  \label{fig:omni_joint}
\end{minipage}
\hspace{0.5cm}
\begin{minipage}[t]{.35\textwidth}
\includegraphics[width=1\textwidth,height=0.12\textheight]{lugs.JPG} \label{fig:lugs_design}
\caption{[1] Lug, [1] Lug-Chain assembly, [3] Circular cross-section of CObRaSO.}
\label{fig:lug} . 
\end{minipage}
\end{figure}
\end{comment}

%\end{itemize}

\subsubsection{2-DOF Chain design}

\begin{figure}[h!]
\centering
\hspace{0cm}
\subfloat[]{\includegraphics[width=0.07\textwidth,height=0.08\textheight]{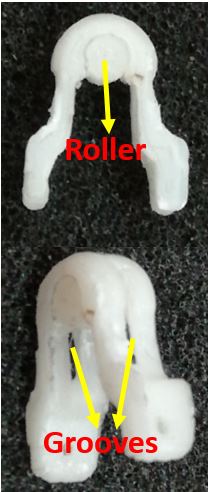} \label{fig:chain_links}}
\hspace{0cm}
\subfloat[]{\includegraphics[width=0.2\textwidth,height=0.08\textheight, angle=-0]{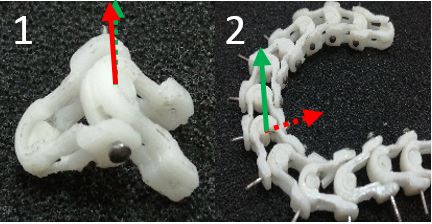}\label{fig:chain_link_rot}} 
\hspace{0cm}
\subfloat[]{\includegraphics[width=0.16\textwidth,height=0.08\textheight, angle=-0]{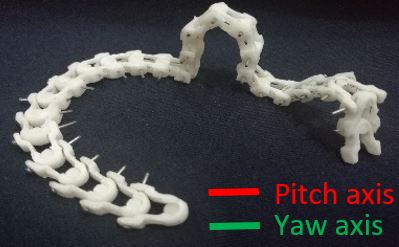}\label{fig:chain_3D}} 
\caption{ (a) A link of the 2-DOF chain link, (b) Rotation about Pitch (left) and Yaw (right) axis, (b) Chain attaining random orientation in a 3D plane owing to its 2-DOF joints. }
\label{fig:chain} 
\end{figure}

In order to comply and crawl with the bent chassis in an Omnidirectional bent configuration of the module, a 2-DOF roller chain has been designed which can rotate about two perpendicular axes and this aids bending about Pitch as well as Yaw axis, which is a unique characteristic of this roller chain. Each chain link comprises of grooves on both of its sides and the successive links are coupled along these grooves. While the rotation of successive chain links about the roller provides the rotation about the Pitch axis, the relative rotation between two links along the grooves facilitates rotation about Yaw axis, as shown in Fig. \ref{fig:chain_links}, \ref{fig:chain_link_rot}. This makes the design and functionality of this chain very unique, and it attains any orientation in the 3-D plane as shown in Fig. \ref{fig:chain_3D}.

\subsubsection{Lug-Chain assembly}
\begin{figure}[h!]
\centering
\hspace{0cm}
\subfloat[]{\includegraphics[width=0.18\textwidth,height=0.08\textheight]{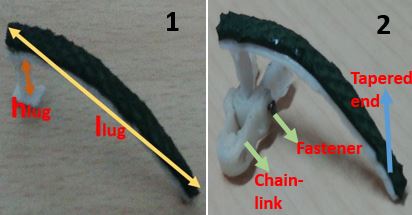} \label{fig:lug}}
\hspace{0cm}
\subfloat[]{\includegraphics[width=0.12\textwidth,height=0.08\textheight]{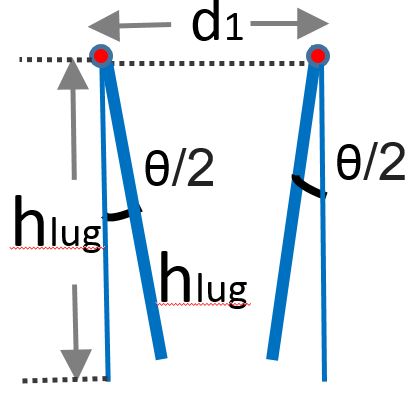} \label{fig:lug_bend_1}}
\hspace{0cm}
\subfloat[]{\includegraphics[width=0.12\textwidth,height=0.08\textheight, angle=-0]{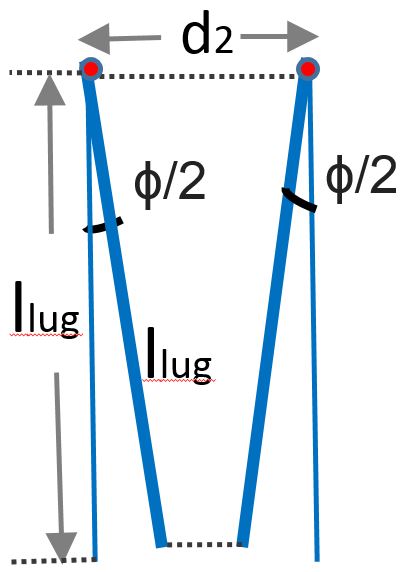}\label{fig:lug_bend_2}} 
\hspace{0cm}
\caption{(a) [1] Nomenclature of the lugs specifications [2] Lug-chain assembly;  Kinematic model of the lugs when lug-chain assembly rotates about (a) Pitch axis and (b) Yaw axis}
\label{fig:chain} 
\end{figure}

The OmniCrawler module possesses an extra degree of freedom, i.e sideways rolling along the Y axis and this enables it to exhibit holonomic motion, which is characterized by the design and arrangement of two identical series of lug-chain assembly. The semi-circular shape of the lugs provides the circular cross-section to the module and the module's ends attain hemispherical shape, as shown in module's prototype in Fig.\ref{fig:module}. The lugs are connected to the chain links via a fastener (Fig. \ref{fig:lug}) and the unique design of chain enables the lug-chain assembly to passively conform with the bent configuration of the module while driving.

The design and arrangement of lugs are determined by the joint angles limit of the module and the diameter of the circular plates. The length of lugs ($l_{lug}$) is optimized to cover the maximum surface of the module. Additionally, the height of lug ($h_{lug}$) is determined by the passage of the lug-chain assembly provided by 3 circular plates and material property of silicone rubber embedded between them. To avoid inter-lug interference when the module bends about Pitch axis ($\theta$) and Yaw axis ($\phi$), the inter-lug separation ($d_{lug}$) is determined by the geometric constraints of the lug and the joint angles limit ($\theta_{max}, \phi_{max}$), as follows.

\begin{align}
\begin{split}
 &d_1 \ge 2h_{lug} \sin\theta_{max}/2\\
 &d_2 \ge 2l_{lug} \sin\phi_{max}/2\\
 &d_{lug} = max(d_1, d_2)
 \end{split}
\label{eqn:lug_sep}
\end{align}
 
For the design parameters listed in Table \ref{table:Design Parameters}, $d_{1}= 5 mm$, $d_{2}= 18 mm$, therefore, $d_{lug}= 18 mm$.

The lugs have been designed with tapered ends (Fig. \ref{fig:lug}) in order to prevent inter-lug interference at the hemispherical ends of the module. Furthermore, the separation between two parallel series of lug-chain assemblies, going through the circular plates passage, is kept as small as possible so that the lugs cover the entire cross-section of the module and lugs provide maximum contact with the surface/terrain while crawling.

\begin{comment}
\begin{figure}[h!]
\centering
%\hspace{-2cm}
\subfloat[]{\includegraphics[width=0.22\textwidth,height=0.09\textheight]{straight_module_1.jpg}\label{fig:straight_module_top} }
\hspace{0cm}
\subfloat[]{\includegraphics[width=0.22\textwidth,height=0.09\textheight]{straight_module_2.jpg}\label{fig:s3traight_module_side} }\\
\subfloat[]{\includegraphics[width=0.22\textwidth,height=0.09\textheight]{bend_module_1.jpg}\label{fig:bend_module_top} }
\hspace{0cm}
\subfloat[]{\includegraphics[width=0.22\textwidth,height=0.09\textheight]{bend_module_2.jpg}\label{fig:bend_module_side} }
\caption{(a) Top view of the straight configuration. (b) Side view of the straight configuration. (c) Top view of the bend configuration. (d) Side view of the Bend configuration }
\label{fig:modules_orientation} 
\end{figure}
\end{comment}

\subsubsection{Silicone Rubber for guiding channel during Omnidirectional bending }

\begin{figure}[h!]
%\subfloat[]{\includegraphics[width=0.1\textwidth,height=0.12\textheight, angle=-0]{mold.jpg}\label{fig:mold}} 
\subfloat[]{\includegraphics[width=0.2\textwidth,height=0.08\textheight]{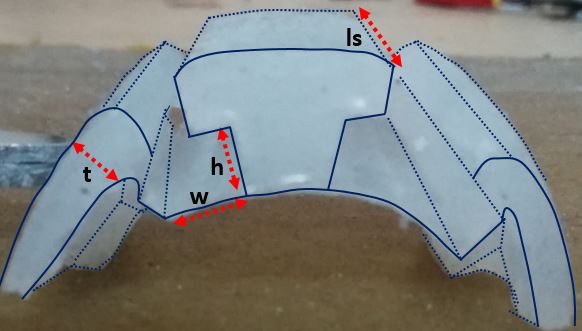} \label{fig:silicone}}
\subfloat[]{\includegraphics[width=0.25\textwidth,height=0.08\textheight]{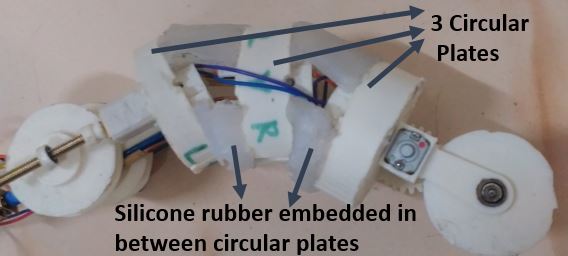}\label{fig:bend_without_chain}}
\caption{(a) Geometry of Silicone rubber, (b) Silicone rubber embedded in between the 3 circular plates provides uniform continuous pathway in-between plates for lug-chain assembly.}
%\label{fig:cross_section} 
\end{figure}

The guiding channel for the two series of lug-chain assembly is provided by 3 circular plates. However, the offset distance in the Omnidirectional joint assembly (incorporated in between these plates) introduces a discontinuity in this pathway (guided channel). As a result of this non-uniform path, the lug-chain assembly tends to get stuck at the edges of the circular plates, which may hinder further movement of chains thereby, leading to either breakage of lug-chain assembly or excessive torque requirement to overcome those edges.
Therefore, soft silicone material of geometry shown in Fig. \ref{fig:silicone} is embedded in between the circular plates to provide continuity in the pathway of lug-chain assembly in any desired directional bending. Additionally, it prevents the robot from mechanical damage on the rugged terrain by distributing the transferred force (between the terrain and robot) evenly over the contact surface. The support provided by the silicone rubber for the chain further prevents the lug-chain assembly from sagging inwards while crawling on the jagged terrain. It also absorbs much of the energy arising from the collision with such terrain thereby, making the module more robust to a highly unstructured surface. 

The Silicone rubber material has been prepared by mixing an equal ratio of two parts (A \& B) of Ecoflex 00-30, Smooth-On Inc. with a 100$\%$ modulus of 69 kPa. The specifications of the mold are determined by following 4 parameters. 1) Characteristics of the silicone rubber, 2) Joint angle limits ($\theta_{max}$,$\phi_{max}$), 3) Dimensions of lug-chain assembly, 4) Shape of edges of the circular plates, as it needs to provide a uniform pathway from soft surface to rigid circular plates. Experimentally selected values of height ($h$) and width ($w$) are listed in Table \ref{table:Design Parameters}.

\begin{figure}[h!]
%\subfloat[]{\includegraphics[width=0.1\textwidth,height=0.12\textheight, angle=-0]{mold.jpg}\label{fig:mold}} 
\subfloat[]{\includegraphics[width=0.22\textwidth,height=0.15\textheight]{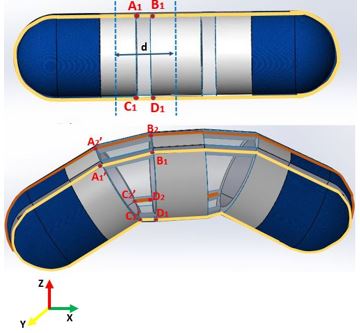} \label{fig:bend_CAD_theta}}
\subfloat[]{\includegraphics[width=0.22\textwidth,height=0.15\textheight]{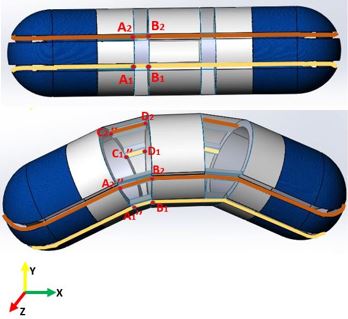}\label{fig:bend_CAD_phi}}
\caption{Module bending about (a) Pitch axis (b) Yaw axis}
\label{fig:bend_CAD} 
\end{figure}

By embedding the soft silicone rubber in between the plates, deformation of silicone is geometrically constrained by the orientation of the circular plates, as shown in FIG. \ref{fig:bend_without_chain}. When the module bends about Pitch axis (Fig. \ref{fig:bend_CAD_theta}), the passive conformation of the silicone rubber with the configuration of the module chassis results in elongation from $A_1B_1$, $A_2B_2$ to $A_1'B_1$, $A_2'B_2$ and bending from $C_1D_1$, $C_2D_2$ to $C_1'D_1$, $C_2'D_2$. Similarly, when module bends about Yaw axis (Fig. \ref{fig:bend_CAD_phi}), rubber elongates from $A_2B_2$, $C_2D_2$ to $A_2''B_2$, $C_2''D_2$ and bends from $A_1B_1$, $C_1D_1$ to $A_1''B_1$, $C_1''D_1$. Since the specified shrinkage of silicone material is $<0.001$ inch/inch, it is assumed to be incompressible and therefore, it bends and bulges out instead of compression. This elongation and bending of the rubber influence the motion behavior of the lug-chain assembly along the guided channel provided by it. While the elongation results in a reduction in the cross-sectional area of the channel ($w$x$h$), which produces more resistance in motion along this path, bending leads to bulging out of rubber which further impedes the motion of the lugs. To determine an optimal trade-off between reduction in cross-sectional area during extension and bulging out during bending, the characteristics of the silicone material were estimated by performing tensile testing on 3 specimens of Silicone rubber with variable thickness values of $t$=2,4 and 6$mm$. The silicone rubber is mathematically modeled as a neo-Hookean material as it exhibits more or less linear or even flattening force-elongation behavior in uniaxial tension, for a finite strain range \cite{c11}. As the sample is pulled uniaxially, the strain in the height ($h$) and the length ($l$) directions are the same. Since, the thin rubber reached breakpoint with less force, based on experimental results, we selected silicone with thickness $t=4mm$ and the rubber was embedded in a pre-extended form which ensures that it does not bulge out when the path length decreases.

%The tensile force-elongation characteristics of three specimen are shown in Fig. \ref{fig:legth_angle}. 
%From the plot shown in Fig. \ref{fig:legth_angle}, the maximum elongation and reduction in length for the joint angle limit ($\theta=32^\circ$) is $14mm$. 

%rom these, the one with $t= 4mm$ demonstrates the optimal trade-off between reduction in cross-sectional area of the available path AB and bulge out along length CD for chain-lug assembly. T 
%The variation along length AB and CD calculated as a function of joint angle (in equation \ref{eqn:chain_len_1}) is further used to estimate the dimensions of the silicone rubber embedded in between the plates. 

\begin{comment}
\begin{figure}[h!]
\centering
\hspace{-1cm}
\subfloat[]{\includegraphics[width=0.2\textwidth,height=0.12\textheight]{chain_links.JPG} \label{fig:chain_links}}
\subfloat[]{\includegraphics[width=0.25\textwidth,height=0.12\textheight, angle=-0]{2D_chain.JPG}\label{fig:chain_link}} 
\caption{(a) The grooves in the chain link enables it to possess an extra degree of freedom, (b) Chain can attain any orientation in the 3D plane, owing to its 2-DOF chain links. }
%\label{fig:cross_section} 
\end{figure}
\end{comment}

\subsection{Tensioner to adjust chain tension in the bent configuration of the module}
The novelty of this module is its ability to crawl in the bent configuration, which is characterized by the unique design of a 2-DOF roller chain. When the module bends, the lug-chain assembly aligns itself with the bent chassis along the path provided by the hybrid structure of rigid (circular plates)-soft (silicone) material. Bending about different axes results in the variation in the path length of the chain-lug assembly, as discussed below.

\begin{figure}[h!]
\centering
\hspace{0cm}
\subfloat[]{\includegraphics[width=0.25\textwidth,height=0.12\textheight]{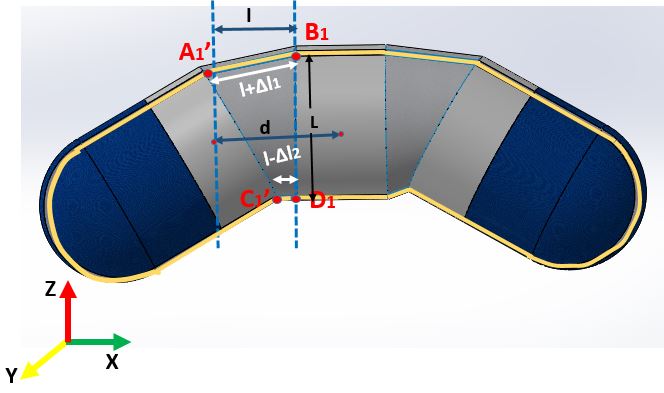}\label{fig:bend_cad_y} }
\hspace{0cm}
\subfloat[]{\includegraphics[width=0.2\textwidth,height=0.12\textheight]{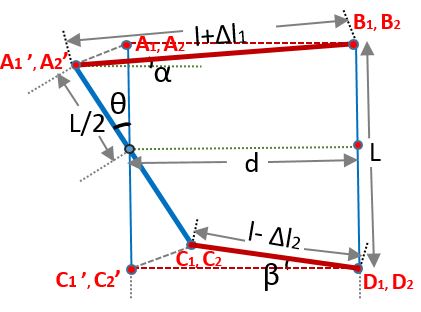}\label{fig:line_y} }
\hspace{0cm}
\subfloat[]{\includegraphics[width=0.25\textwidth,height=0.12\textheight]{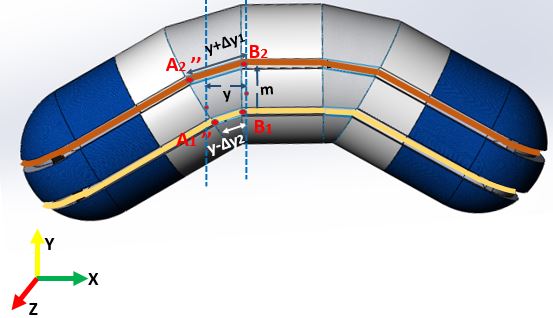}\label{fig:bend_cad_z} }
\hspace{0cm}
\subfloat[]{\includegraphics[width=0.2\textwidth,height=0.12\textheight]{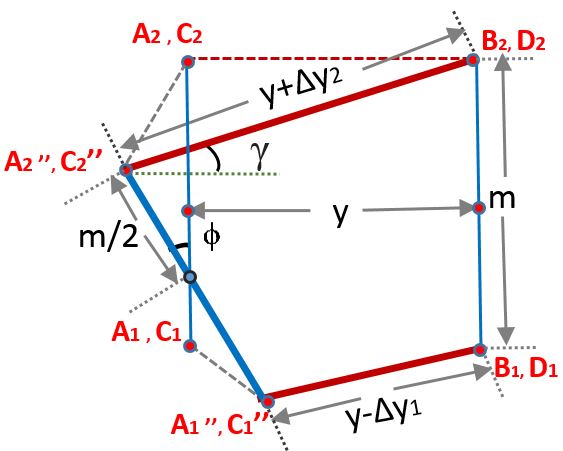}\label{fig:line_z} }
\caption{ (a) Bending about Y-axis (Pitch angle $\theta$), (b) Kinematic model showing the geometry of variation of the lug-chain assembly path length when module bends with angle $\theta$, (c) Bending about Z-axis (Pitch angle $\phi$)(d) Kinematic model showing the geometry of variation of the lug-chain assembly path length when module bends with angle $\phi$.}
\label{fig:chain_tension_specs}
\end{figure}

\begin{figure}[h!]
\hspace{0cm}
\subfloat[]{\includegraphics[width=0.24\textwidth,height=0.14\textheight, angle=-0]{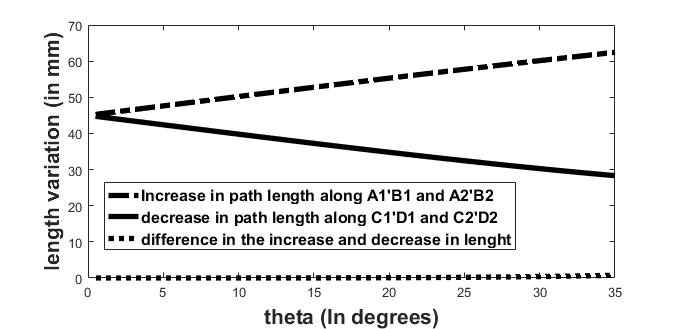}\label{fig:length_theta}} 
\subfloat[]{\includegraphics[width=0.24\textwidth,height=0.14\textheight, angle=-0]{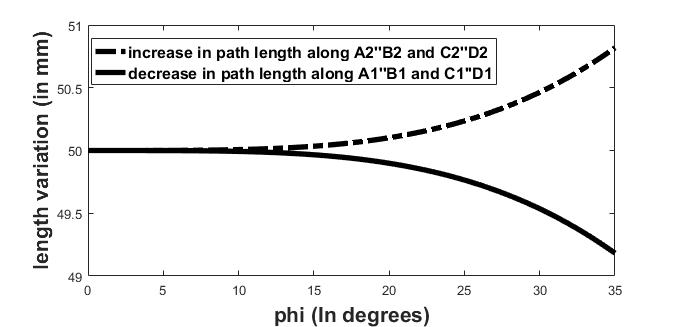}\label{fig:length_phi}} 
\caption{ Plots showing variation in path length as a function of (a) $\theta$, (b) $\phi$ }
\label{fig:length_angle} 
\end{figure}

\textit{1) Bending about Pitch axis ($\theta$)} $\colon$
The bending of module about Pitch axis (Fig. \ref{fig:bend_cad_y}) results in an increase and decrease in the Silicone path length along from $A_1B_1$, $C_1D_1$ to $A_1'B_1$, $C_1'D_1$, respectively, as illustrated by the kinematic model in Fig. \ref{fig:line_y}. This variation in the Silicon path length of the lug-chain assembly is determined by the geometric constraints of the module w.r.t joint angle ($\theta$), as following. 

\begin{align}
\begin{split}
 &\tan \alpha = \frac{L(1-\cos\theta)}{2d+L\sin\theta}\\
 &\tan \beta = \frac{L(1-\cos\theta)}{2d-L\sin\theta}\\
 &L_{A_{1}'B_{1}} = L_{A_{2}'B_{2}} = l-\Delta l_1 = \frac{L(1-\cos\theta)}{2 \sin\alpha}\\
 &L_{C_{1}'D_{1}} = L_{C_{2}'D_{2}} = l-\Delta l_2 = \frac{L(1-\cos\theta)}{2 \sin\beta}
 \end{split}
\label{eqn:random}
\end{align}
The plot in Fig. \ref{fig:length_theta} shows the variation in the length along path $L_{A_{1}'B_{1}}$ and $L_{C_{1}'D_{1}}$ w.r.t $\theta$. From the plot, the difference in variation in both lengths is negligible for the range of angles ($ 0 \le  \theta \le 25^ \circ$), hence the resultant change in length from path $A_1B_1D_1C_1$ to $A_1'B_1D_1C_1'$ is almost negligible, 
$\therefore  \Delta l_1 - \Delta l_2 = 0 $

As a result, the geometry of the chain links enables it keep the chain intact and maintain tension even during bent configuration.

\textit{2) Bending about Yaw axis ($\phi$) $\colon$}
Secondly, when module bends along a plane other than X-Z plane, chain links involve rotation about the Yaw axis which leads to passive conformation of the
chain along the grooves in the chain-link. This leads to a variation in the path length of the chain which influence the tension of the track. Therefore, the
tension in the chains needs to be adjusted to maintain optimal tension for crawling in the bent configuration. These are further illustrated mathematically using the kinematic model shown in Fig. \ref{fig:line_z}.The extension and reduction in the path length  w.r.t joint angle ($\phi$) about Z axis, are given by the following relations.

\begin{align}
\begin{split}
& \tan\gamma =\frac{2(1-\cos\phi)}{\cos\phi+\sin\phi}\\
&\Delta y_1 = \Delta y_2 = \frac{m(1-\cos\phi)}{2\sin\gamma}\\
&L_{A_{1}''B_1} = L_{C_{1}''D_1} = y-\Delta y_1\\ 
&L_{A_{2}''B_2} = L_{C_{2}''D_2} =  y+\Delta y_2\\
&L_{A_1''B_1D_1C_1''} = L_{A_1B_1D_1C_1}-2\Delta y_1\\
&L_{A_2''B_2D_2C_2''} = L_{A_2B_2D_2C_2}+2\Delta y_2\\
\end{split}
\label{eqn:chain_len_2}
\end{align}

The plot in Fig. \ref{fig:length_phi} shows the variation in the path length along $A_{1}''B_{1}$ and $C_{1}'D_{1}$ w.r.t joint angle ($\phi$). Since the two path lengths ($L_{A_{2}''B_{2}}, L_{C_{2}''D_{2}}$) of one of the chains ($A_2 B_2 D_2 C_2$) increases, the resultant length of the chain increases by $2\Delta y_2$. Similarly, the length of the chain along path ($A_1 B_1 D_1 C_1$) reduces by $2\Delta y_1$.

\begin{comment}
\begin{figure}[h!]
\centering
{\includegraphics[width=0.4\textwidth,height=0.12\textheight]{tensioner.jpg}} 
\caption{Figure shows the linear screw actuator mechanism used to adjust the chain tension while crawling in bent position.}
\label{fig:tension}
\end{figure}
\end{comment}

The chain tension is adjusted by means of a mechanism using linear screw actuators. The actuator adjusts the position of one of the sprockets corresponding to each of the 2 chains, with the aid of a slider attached to the center of these sprockets. The linear screw actuator slides the position of the sprockets back and forth which consequently adjusts the tension of the corresponding chain. The arrangement of linear screw actuator (tensioner), sprocket and the slider are shown in CAD model Fig. \ref{fig:CAD}.

\section{Locomotion modes  }
\label{Sec:locomotion}

The versatility of CObRaSO enables its integration in several categories of robotic platforms, which enables them to exhibit various locomotion modes to navigate the dynamically varying environment. Path planning in such an environment can be achieved by transforming among various possible locomotion modes, which are further decomposed into simple motion primitives, such as Crawl, bend, roll, steer, etc. These motion primitives are characterized by the knowledge of the environment and the kinematics of the robot as described in subsequent sections and summarized in Table \ref{table:primitves}.

\subsection{Crawler Locomotion Mode}

A combination of longitudinal Crawling motion and the sideways rolling motion enables holonomic motion of the module, which is the basic characteristic of an OmniCrawler module \cite{c15}, as shown in Fig. \ref{fig:basic_perform}. With the controlled actuation of the joints of CObRaSO about Pitch and Yaw axes, it can crawl while actively complying with the geometry of the obstacle in a 3-D unstructured environment and can overcome the obstacle either by steering around it or climbing over it. 
For a dynamic terrain, the module's joint angles about yaw axis ($\phi_i$), can be determined using Inverse kinematics (IK).

\begin{align}
\begin{split}
\phi_i, \theta_i = IK_{3D}(ObstaclePose,ObstacleGeometry)
\end{split}
\label{eqn:crawl_3D}
\end{align}

\subsubsection{Steering and Obstacle Aided Motion       (Crawling + Yaw Joints actuation)}
\begin{figure}[h!]   
\centering
\subfloat[]{\includegraphics[width=0.25\textwidth,height=0.1\textheight, angle=-0]{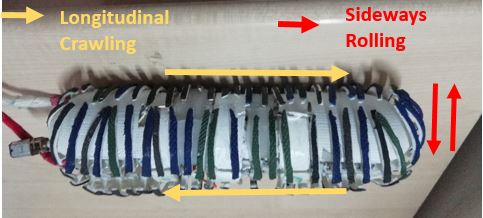}\label{fig:basic_perform}}
\subfloat[]{\includegraphics[width=0.2\textwidth,height=0.1\textheight, angle=-0]{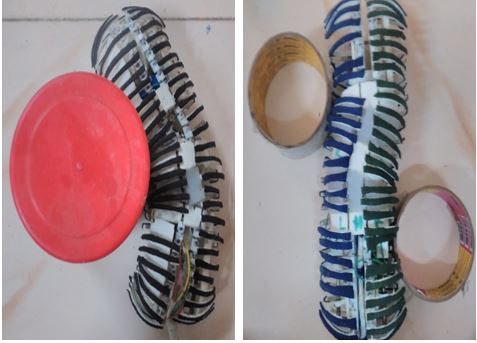}\label{fig:steering_2}}

\caption{(a) Basic locomotion mode of an OmniCrawler module involve longitudinal crawling and sideways rolling mtion, (b) Experiments demonstrating CObRaSO complies with the curvature of the obstacle executing obstacle aided motion. }
\end{figure}

The motion of two parallel longitudinal series of lugs-chain assembly over the surface of the module facilitates surface actuation of the module. While crawling, the surface actuation property of the module exploits external obstacles and terrain irregularities in a cluttered environment and achieves obstacle-aided motion for forward motion \cite{c16}, as illustrated in \ref{fig:steering_2}. This property is very useful for snake robot motion in confined and cluttered spaces.

\subsubsection{Navigating 3D unstructured environment (Crawling + Sideways Rolling + Yaw $\&$ Pitch Joints actuation)}
\label{conform}

\begin{figure}[h!]   
\subfloat[]{\includegraphics[width=0.25\textwidth,height=0.08\textheight, angle=-0]{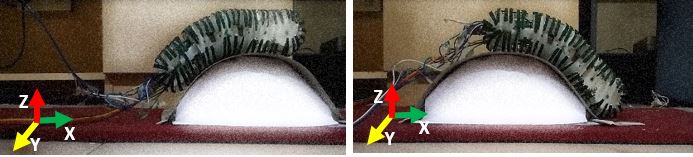}\label{fig:climb_obs}}
\subfloat[]{\includegraphics[width=0.25\textwidth,height=0.08\textheight, angle=-0]{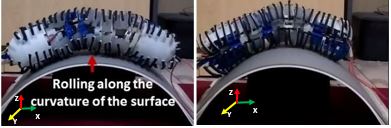}\label{fig:uneven_terrain}}

\caption{Experiments demonstrating the robot complying with the surface curvature as it undergoes (a) Crawling while traversing in forward direction., (b) sideways rolling motion along the curvature of the surface.}
\end{figure}

The module can traverse dynamically varying 3D environment by combining longitudinal crawling or sideways rolling motion with the Omnidirectional bending of the module. While the combination of Omnidirectional bending with Crawling facilitates complying with the unevenness of surface while moving forward, as shown in Fig. \ref{fig:climb_obs}, its combination with sideways rolling achieves compliance on an uneven surface during sideways rolling motion, as shown in Fig. \ref{fig:uneven_terrain}. Therefore, a combination of all three motion primitives (Bend, Crawl, Roll) achieves navigation in a highly unstructured 3D environment. This can be effectively used for Spiral Stair Climbing Robots.

\begin{comment}

\begin{figure}[h!]
\centering
%\hspace{-2cm}
\subfloat[]{\includegraphics[width=0.15\textwidth,height=0.15\textheight,angle=-0] {turning.jpg}\label{fig:turning_obs}} 
\hspace{1cm}
\subfloat[]{\includegraphics[width=0.15\textwidth,height=0.15\textheight, angle=-0]{turning.jpg}\label{fig:turning_obs}} 

\caption{(a) Module avoids obstacle by turning around the obstacle..(ADD PICs OF VARIOUS STEPS INVOLVED IN THIS EXP).}
\end{figure}
\end{comment}

\subsection{Legged locomotion Mode}
The joints actuation within the CObRaSO module has the potential benefits of exhibiting legged locomotion in a various robotic platform such as a Quadruped. This characteristic can be further used in several applications as a manipulator and a gripper which enables it to interchangeably exhibit manipulation and locomotion capabilities by switching its modes these.

\subsection{Wheeled Locomotion Mode}
An OmniCrawler is an extension of the Omniball from a sphere to a cylinder with spherical ends. When the module crawls with one of the hemispherical ends in contact with the surface, the motion of lug-chain assembly across the surface of the module produces a wheeled locomotion characteristic similar to an Omni-wheel \cite{c9}.

The legged and wheel locomotion modes of CObRaSO are further demonstrated in Section \ref{Sec:applications} with the aid of a simulation and experimental results of a Quadruped.
%\section{Joint Analysis}

%The joint analysis of the proposed design is presented here.

\begin{table}[t] 
\centering
\caption{Locomotion modes decomposed into several motion primitives}
\begin{tabular}{|p{3cm}|p{5cm}|}
\hline
\textbf{Locomotion Mode} & \textbf{Motion Primitive}  \\
\hline
Omnidirectional motion  & Longitudinal Crawling motion + Sideways Rolling motion \\

\hline
Steering Mode or Compliant Crawling Mode & Joints control + Chains tension adjustment + Crawling motion \\

\hline
Compliant Rolling Mode &  Joints control + Sideways Rolling motion \\

\hline
Legged Locomotion & Joints control \\

\hline

\end{tabular}
\label{table:primitves}
\end{table}

\begin{comment}
\begin{figure*}[htp!]   
\hspace{-2cm}
\subfloat[]{\includegraphics[width=0.1\textwidth,height=0.08\textheight, angle=-0]{quad_forward.pdf}\label{fig:quad_Straight}}\\
\hspace{-2cm}
\subfloat[]{\includegraphics[width=0.1\textwidth,height=0.08\textheight, angle=-0]{quad_sideways.pdf}\label{fig:quad_sideways}}\\
\hspace{-2cm}
\subfloat[]{\includegraphics[width=0.1\textwidth,height=0.08\textheight, angle=-0]{quad_tilt.pdf}\label{fig:quad_tilt}}\\
\subfloat[]{\includegraphics[width=1\textwidth,height=0.08\textheight,angle=-0] {gaits_in_quad.pdf}\label{fig:quad_gaits}} 
\caption{ In Quadruped configuration, (a) robot crawls forward while complying dynamically with the unevenness of surface and climbing at a height 9cm. (b) robot rolls sideways on uneven surface, (c) omni-wheel locomotion mode on uneven terrain. (d) To navigate under confined spaces, the robot switches from legged to crawler model of locomotion. } \label{fig:quad_adams}
\end{figure*}

\end{comment}

\section{ Applications and Experiments}
\label{Sec:applications}
To validate the proof of concept of modular design of CObRaSO and showcase its versatility in various categories of robotic platforms, we developed prototypes of Quadruped, an In-Pipe climbing robot, and a snake robot by cascading different numbers of modules. Subsequently, various experiments and simulations were conducted in a dynamic environment to demonstrate the adaptability and reconfigurability of each one of them. Also, the simulations were carried out in ADAMS, a multi-body dynamics simulator with the lumped model of each module, where each CObRaSO module is represented as a cascade of omni-wheels. The motion primitives were generated intuitively for each robot platform for planning motions in different environments, such as overcoming the obstacle (for a Snake), negotiating 3D pipe bends (for an In-Pipe climber), and tight spaces (for a Quadruped). These set of motion primitives were commanded by an operator manually.

\subsection{Snake Robot}
\label{Sec:snake}

\begin{figure}[htp!]   
\centering
\subfloat[]{\includegraphics[width=0.22\textwidth,height=0.06\textheight]{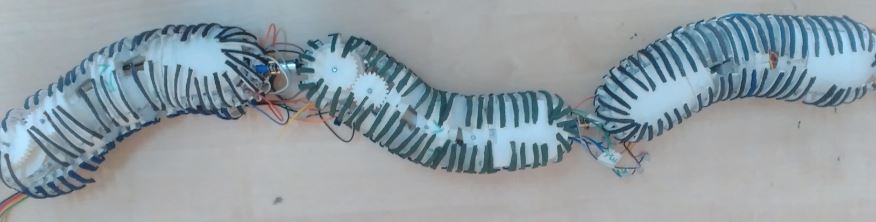}\label{fig:snake}}
\centering
\subfloat[]{\includegraphics[width=0.22\textwidth,height=0.06\textheight]{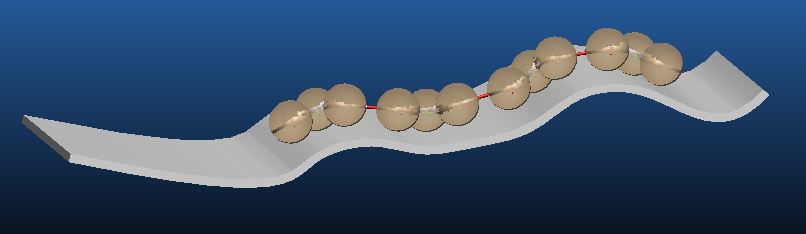}\label{fig:snake_simulation}}\\

\caption{ (a) A Snake robot realized using Kinematic chain of 3 modules, (b) A montage of simulation showing Snake robot  traversing uneven terrain.}
\end{figure}

The snake robot, shown in Fig. \ref{fig:snake}, is realized with a kinematic chain of CObRaSO modules. The potential advantage of this design using CObRaSO module lies in its simple modes of locomotion. For instance, the forward crawling and sideways rolling motion can be performed by the Omnidirectional motion capability of each of the OmniCrawler module, assuming the links connecting the adjacent modules are heavy enough to roll each module using external rolling motors. This eases the locomotion of these robots which would otherwise need complex gaits (slithering or side-winding gaits) to perform even forward or sideways motion. The set of motion primitives for a snake robot to negotiate an uneven terrain is represented in Algorithm \ref{alg:algo} and one of the montages corresponding to the simulation results is shown in Fig. \ref{fig:snake_simulation}. 

%%%%%%%%%%%%%%%%%%%%%%%%%%

\begin{comment}
\begin{algorithm}
  \caption{Snake robot traversing in a known unstructured environment with optimal locomotion mode}\label{alg:snake_algo}
  \begin{algorithmic}[1]
    \Procedure{SensorData}{$N, n, ObstacleGeometry, ObstaclePose$}\\
  	%\State $\theta_p\gets g(\theta_{ij}) $
		%\State $\theta_r\gets g(\theta_{ij}) $
		%\State $H_{est}\gets f(\theta_p,\theta_r) $ 				\Comment{Estimate height of the obstacle}
        %H,\theta_p,\theta_r 
      \For{\texttt{< i=N>1, j=n >}} 
         \If {\texttt{<flat terrain without obstacles>}}
        \State  $\theta_{ij}, \phi_{ij}\gets 0 $
            \State $Motion \gets CrawlOrRoll()$
         \EndIf 

        \If{\texttt{<flat surface with obstacles>}}
          \State  $\phi_{ij}\gets IK_{Crawl2D}(ObstacleGeometry, ObstaclePose) $
        \State  $\theta_{ij}\gets 0 $
              \State $Motion \gets Crawl()$
         \EndIf 
 		
        \If{\texttt{<3D uneven surface>}}
           	  \State  $\phi_{ij}, \theta_{ij} \gets IK_{Crawl3D}(ObstacleGeometry, ObstaclePose) $ 
              \State  $l_{mn} \gets Adjust Chain Tension(\phi_{ij}, \theta_{ij})$
        	  \State  $Motion \gets  CrawlAndRoll()$
         \EndIf 
       
        \EndFor

\EndProcedure
  \end{algorithmic}
\end{algorithm}
%%%%%%%%%%%%%%%%%%%%%%%%%%

\end{comment}

%%%%%%%%%%%%%%%%%%%%%%%%%%%%%%%
\begin{algorithm}
  \caption{Locomotion modes for various robots}\label{alg:algo}
  \begin{algorithmic}[1]
    \Procedure{Input}{$Sensor Data, KinematicModel$ }
    
      \For{\texttt{<Snake robot for 3D uneven terrain>}}  \Comment{Fig. \ref{fig:snake_simulation}}
            \State  $CompliantCrawlingMode()$
      \EndFor

         \For{\texttt{<Pipe climbing robot to navigate 
         $90^\circ$ bend>}} \Comment{Fig.\ref{fig:pipe_1}-\ref{fig:pipe_6}}
             \State $AdjustSEAstiffness() $
			\State  $Rolling Motion ()$  \Comment{Minimal energy posture}
              \State  $CompliantCrawlingMode()$ \Comment{to negotiate bend}
      \EndFor

          \For{\texttt{<Quadruped to navigate confined spaces>}} \Comment{Fig.\ref{fig:quad_trans}}
             \State $TransformToCrawler() $ \Comment{from initial configuration}
			\State  $Crawling()$  
      \EndFor

\EndProcedure
  \end{algorithmic}
\end{algorithm}

\subsection{In-Pipe Climbing Robot}

\begin{figure}[htp!]   
\centering
\subfloat[]{\includegraphics[width=0.25\textwidth,height=0.08\textheight,angle=-0] {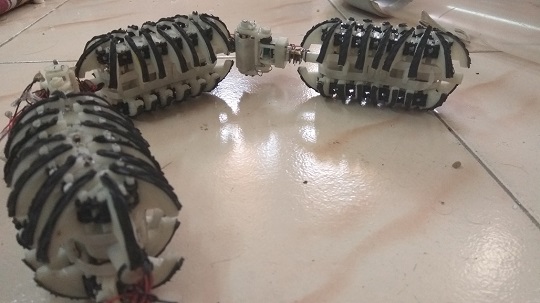}\label{fig:pipe_robot}} 
\centering
\subfloat[]{\includegraphics[width=0.2\textwidth,height=0.08\textheight,angle=-0] {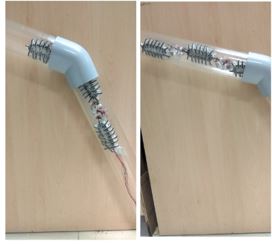}\label{fig:pipe_45}}

\caption{(a) An In-pipeclimbing robot realised with a kinematic chain of 3 modules, (b) Montages of robot demonstrating experiment in a $45^\circ$ bent elbow }

\end{figure}

An in-Pipeline robot, shown in Fig. \ref{fig:pipe_robot}, is realized with a kinematic chain of 3 modules interconnected by 2 Series elastic actuators (SEA) joints \cite{c14}. In straight pipes, all 3 modules are aligned in-line with the pipe and are driven synchronously to propagate in forward/backward direction. The stiffness of the SEAs is adjusted according to the curvature of the pipe which provides the necessary clamping force to overcome robot’s own body weight and facilitate slip free driving motion. The modular configuration of the robot handles pipe diameter variations and sideways rolling motion enables it to overcome bends with minimal energy posture \cite{c10}. The compliance further facilitates locomotion in sharp bends in small diameter pipes by complying with the curvature of the bend.

\begin{figure*}[h!]
\hspace{-1cm}
 \subfloat[]{\includegraphics[width=0.2\textwidth,height=0.08\textheight,angle=-0] {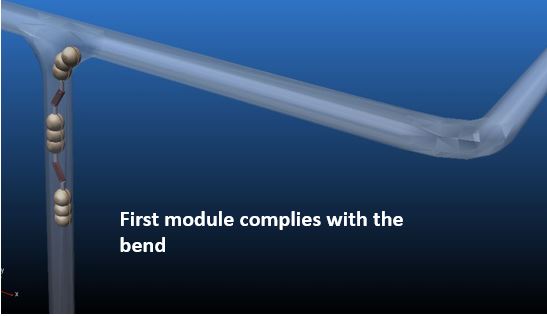}\label{fig:pipe_1}}
\hspace{-0.5cm}
\subfloat[]
{\includegraphics[width=0.2\textwidth,height=0.08\textheight,angle=-0] {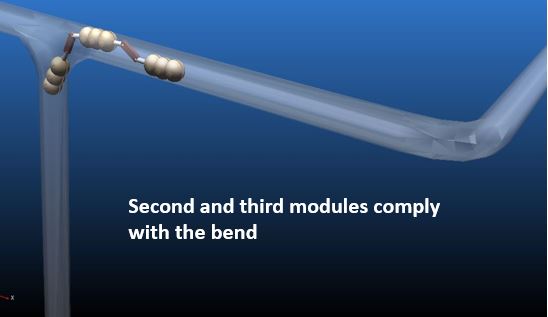}\label{fig:pipe_2}} 
\hspace{-0.5cm}
\subfloat[]{\includegraphics[width=0.2\textwidth,height=0.08\textheight,angle=-0] {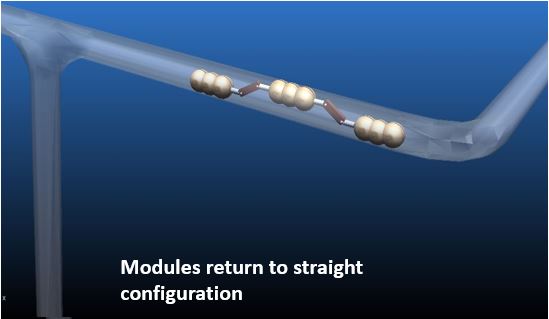}\label{fig:pipe_3}}
\hspace{-0.5cm}
\subfloat[]{\includegraphics[width=0.2\textwidth,height=0.08\textheight,angle=-0] {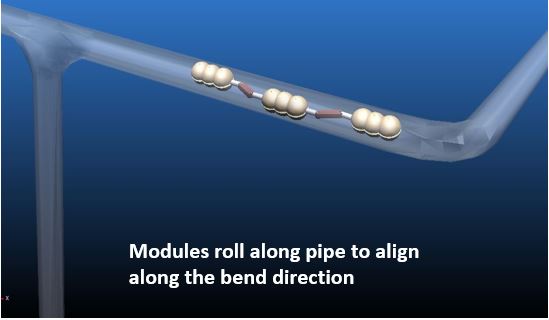}\label{fig:pipe_4}}
\hspace{-0.5cm}
\subfloat[]{\includegraphics[width=0.2\textwidth,height=0.08\textheight,angle=-0] {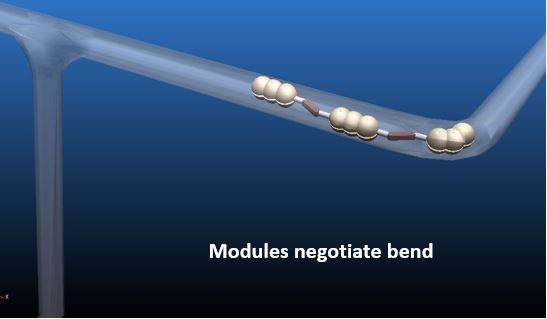}\label{fig:pipe_5}}
\subfloat[]{\includegraphics[width=0.2\textwidth,height=0.08\textheight,angle=-0] {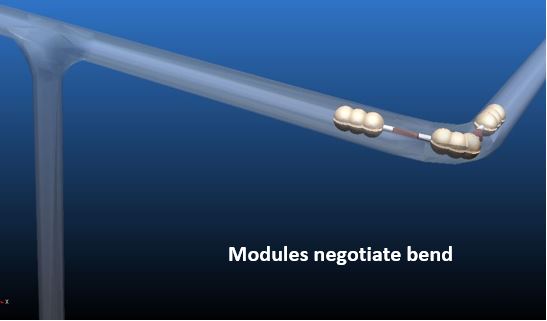}\label{fig:pipe_6}}

\subfloat[]
{\includegraphics[width=1\textwidth,height=0.08\textheight,angle=-0] {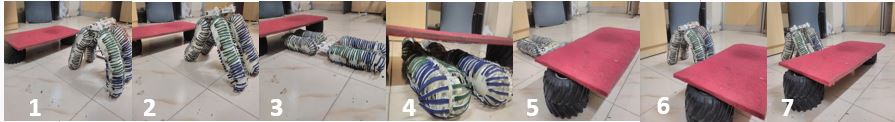}\label{fig:quad_trans}}
\hspace{0cm}
\caption{(a)-(f) Simulation results demonstrating traversal in demonstrating locomotion of robot in sharp 90$^\circ$ bend. (g)Montages showing transformation of Qudruped locomotion modes from legged to crawler robot to navigate confined spaces}\label{fig:pipe_quad}
\end{figure*}

In out previous In-Pipe climbing robot \textit{COCrIP} \cite{c10}, planar 1-DOF joints were incorporated within each OmniCrawler module. Few montages of COCrIP robot demonstrating negotiation of a sharp 45$^\circ$ in a vertical pipe is shown in Fig.\ref{fig:pipe_45}. More experiments can be viewed here (https://youtu.be/Esd47RzzpOU). 
However, the module with 1-DOF planar joints limits its applications to negotiate planar bends aligned in the direction of the module's joint and it becomes difficult to negotiate non-planar pipe bends. With the proposed design of CObRaSO module, the various sequence of motion primitives to negotiate non-planar 90$^\circ$ bend is illustrated in Algorithm \ref{alg:algo} and shown in Fig.\ref{fig:pipe_1}-\ref{fig:pipe_6}.

%%%%%%%%%%%%%%%%%%%%%%%%%%%%%%%
\begin{comment}
\begin{algorithm}
  \caption{An In-Pipe climbing robot traversal in a known 3 dimensional Pipe environment with minimal energy locomotion mode}\label{alg:algo_pipe}
  \begin{algorithmic}[1]
    \Procedure{SensorData}{$N, n, Diameter, BendDirection, BendAngle, BendCurvature$ }\\
      \For{\texttt{<i=N=3, j=n>}} 
         \If{\texttt{<straight pipe>}}
            \State  $\phi_{ij}, \phi_{ij}\gets 0 $
            \State  $SEA \gets Adjuststiffness (Diameter) $
            \State  $Motion \gets Crawl()$
         \EndIf

         \If{\texttt{<Pipes with T junction>}}
             \State  $SEA \gets Adjuststiffness(Dia, Curv) $
			\State  $Align \gets RollAlongPipe(BendDirection) $
              \State  $Motion \gets Crawl()$
         \EndIf

         \If{\texttt{<Pipes with one sided turn>}}
			\State  $SEA \gets Adjuststiffness(Diameter) 
            $\State  $ Align \gets RollAlongPipe(BendDirection) $
            \State  $SEA \gets Adjuststiffness(Curv) $
            \State  $\phi_{ij}, \theta_{ij}\gets IK_{pipebend}(Curv) $
            \State  $l_{mn} \gets Adjust Chain Tension(\phi_{ij}, \theta_{ij}), 								\theta_{R_{i}})$
            \State  $Motion \gets Crawl()$
         \EndIf

        \EndFor

\EndProcedure
  \end{algorithmic}
\end{algorithm}
\end{comment}

%The motion primitives to overcome sharp planar 90$^\circ$ bend are demonstrated in Fig. 

\subsection{Quadruped}

\begin{figure}[h!]
\centering
\subfloat[]
{\includegraphics[width=0.45\textwidth,height=0.08\textheight,angle=-0] {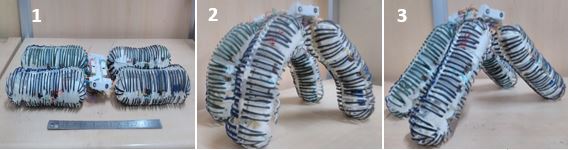}\label{fig:quad_bot}}\\
\subfloat[]
{\includegraphics[width=0.45\textwidth,height=0.06\textheight,angle=-0] {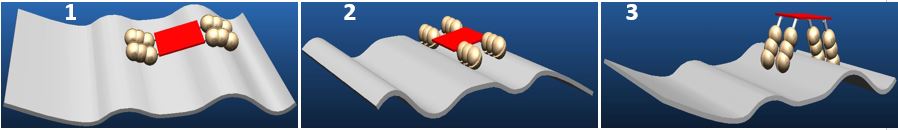}\label{fig:quad_sim}}
\caption{Quadruped in [1] Crawler configuration, [2] Legged configuration, [3] Wheeled configuration; (b) Montages showing simulation results where Quadruped crawls forward while complying dynamically with the unevenness of surface by [1] complying with the surface, [2] sideways rolling, [3] Wheeled mode}
\label{fig:quad}
\end{figure}

\begin{figure}
\subfloat[]
{\includegraphics[width=0.3\textwidth,height=0.08\textheight,angle=-0] {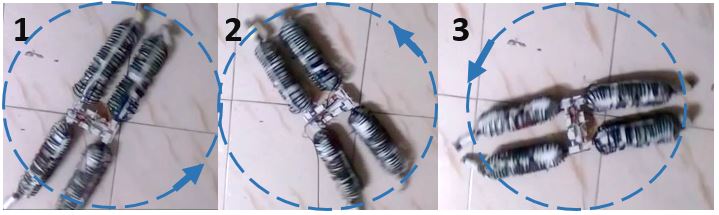}\label{fig:quad_holo}}
\subfloat[]{\includegraphics[width=0.2\textwidth,height=0.08\textheight,angle=-0] {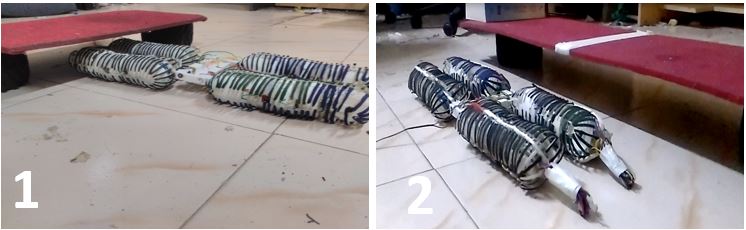}\label{fig:quad_under}}
\caption{ (a) Quadruped performing circular motion by differential rolling motion of its modules; (b) Quadruped navigating under constrained surface either by }
\end{figure}

The proposed module is configured as each of the 4 legs in Quadruped robot and its versatility enables the robot to attain various configurations to perform different locomotion modes, as shown in Fig. \ref{fig:quad_bot}. The simulation results demonstrating the motion of each of these modes on an uneven surface are shown in Fig. \ref{fig:quad_sim}. The differential sideways rolling motion of each of these modules perform the holonomic circular motion, as shown in Fig. \ref{fig:quad_holo}. With the initial configuration as legged (or wheeled robot), the set of motion primitives corresponding to an instance where Quadruped switches its locomotion modes from legged to crawler mode to go under tight spaces is described in Algorithm \ref{alg:algo} and further demonstrated in Fig. \ref{fig:quad_trans}.

\section{Conclusion and Future Work}
In this paper, we discuss the design of a novel Omnidirectional bendable OmniCrawler module- CObRaSO. The ability of the module to crawl using a single pair of the chain while bending in any desired direction demonstrates the uniqueness of this design. This helps to achieve high maneuverability and adaptability on an uneven surface. Its unique modular design facilitates integration in several robotic platforms and also enhances their locomotion capabilities. However, all the motion primitives to perform a locomotion mode in several robots were commanded manually by an operator. Therefore, our future work involves sensor integration within the module to reconfigure itself into different robots and perform the optimal locomotion mode to navigate dynamically varying complex terrain. Additionally, we plan to extend the applicability of this module for a Humanoid, Gripper and Spiral stair climbing robot. 

%For a given pipe environment, a set of optimal spring stiffness values were calculated to quasi-statically balance the robot in straight as well as bend pipes. Afterwards, the limiting value of friction coefficient which could be climbed by the robot without slippage, was calculated. However, the compliant active joint has 1 DOF, which restricts the bending in one direction. This requires robot to attain a configuration corresponding to that direction. Therefore, our future work would focus on modification of the design to bend each module in all directions. Furthermore, a torque control strategy for SEA to comply with diameter and friction coefficient variation would be implemented.

\end{document}